\documentclass[10pt,twocolumn,letterpaper]{article}

\usepackage{cvpr}
\usepackage{times}
\usepackage{epsfig}
\usepackage{graphicx}
\usepackage{amsmath}
\usepackage{amssymb}
\usepackage{subcaption}
\usepackage{color}

\usepackage[pagebackref=true,breaklinks=true,letterpaper=true,colorlinks,bookmarks=false]{hyperref}

\cvprfinalcopy % *** Uncomment this line for the final submission

\def\cvprPaperID{1163} % *** Enter the CVPR Paper ID here

\ifcvprfinal\fi

\begin{document}

%%%%%%%%% TITLE
\title{Residual Attention Network for Image Classification}

\author{Fei Wang$^1$, Mengqing Jiang$^2$, Chen Qian$^1$, Shuo Yang$^3$, Cheng Li$^1$, \\
Honggang Zhang$^4$, Xiaogang Wang$^3$, Xiaoou Tang$^3$ \\
$^1$SenseTime Group Limited, $^2$Tsinghua University,\\
 $^3$The Chinese University of Hong Kong, $^4$Beijing University of Posts and Telecommunications \\
\tt\small$^1$\{wangfei, qianchen, chengli\}@sensetime.com, $^2$jmq14@mails.tsinghua.edu.cn\\
\tt\small$^3$\{ys014, xtang\}@ie.cuhk.edu.hk, xgwang@ee.cuhk.edu.hk, $^4$zhhg@bupt.edu.cn
%\and
}

\maketitle
\thispagestyle{empty}

%%%%%%%%% ABSTRACT
\begin{abstract}
%Mixed nature of human attention has been proposed in the literature of biology and been applied to sequential learning task using RNN and LSTM.
In this work, we propose ``Residual Attention Network", a convolutional neural network using attention mechanism which can incorporate with state-of-art feed forward network architecture in an end-to-end training fashion.
Our Residual Attention Network is built by stacking Attention Modules which generate attention-aware features. The attention-aware features from different modules change adaptively as layers going deeper. Inside each Attention Module, bottom-up top-down feedforward structure is used to unfold the feedforward and feedback attention process into a single feedforward process. Importantly, we propose attention residual learning to train very deep Residual Attention Networks which can be easily scaled up to hundreds of layers.

Extensive analyses are conducted on CIFAR-10 and CIFAR-100 datasets to verify the effectiveness of every module mentioned above. Our Residual Attention Network achieves state-of-the-art object recognition performance on three benchmark datasets including CIFAR-10 (3.90\% error), CIFAR-100 (20.45\% error) and ImageNet (4.8\% single model and single crop, top-5 error). Note that, our method achieves \textbf{0.6\%} top-1 accuracy improvement with \textbf{46\%} trunk depth and \textbf{69\%} forward FLOPs comparing to ResNet-200. The experiment also demonstrates that our network is robust against noisy labels.
\end{abstract}

%%%%%%%%% BODY TEXT
\section{Introduction}

\begin{figure*}
\begin{center}
%\fbox{\rule{0pt}{2in} \rule{.9\linewidth}{0pt}}
\includegraphics[width=1\linewidth]{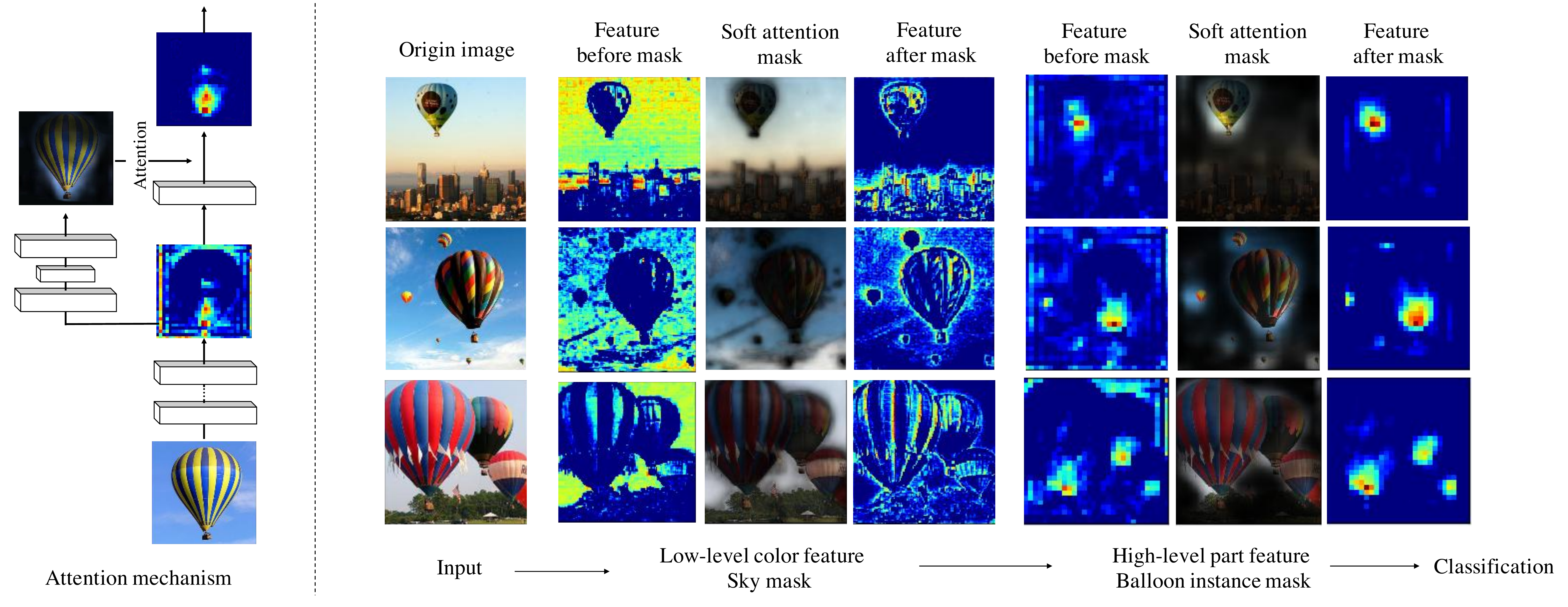}
\end{center}
   \caption{\textbf{Left:} an example shows the interaction between features and attention masks. \textbf{Right:} example images illustrating that different features have different corresponding attention masks in our network. The sky mask diminishes low-level background blue color features. The balloon instance mask highlights high-level balloon bottom part features.}
\label{fig:motivation}
\end{figure*}

Not only a friendly face but also red color will draw our attention. The mixed nature of attention has been studied extensively in the previous literatures~\cite{walther2002attentional, itti2001computational,mnih2014recurrent,zhao2016diversified}. Attention not only serves to select a focused location but also enhances different representations of objects at that location. Previous works formulate attention drift as a sequential process to capture different attended aspects. However, as far as we know, no attention mechanism has been applied to feedforward network structure to achieve state-of-art results in image classification task. Recent advances of image classification focus on training feedforward convolutional neural networks using ``very deep" structure~\cite{simonyan2014very,szegedy2015going,resnet2016}.

Inspired by the attention mechanism and recent advances in the deep neural network, we propose Residual Attention Network, a convolutional network that adopts mixed attention mechanism in ``very deep" structure. The Residual Attention Network is composed of multiple Attention Modules which generate attention-aware features. The attention-aware features from different modules change adaptively as layers going deeper.

Apart from more discriminative feature representation brought by the attention mechanism, our model also exhibits following appealing properties:

\noindent
(1) Increasing Attention Modules lead to consistent performance improvement, as different types of attention are captured extensively. Fig.\ref{fig:motivation} shows an example of different types of attentions for a hot air balloon image. The sky attention mask diminishes background responses while the balloon instance mask highlighting the bottom part of the balloon.

\noindent
(2) It is able to incorporate with state-of-the-art deep network structures in an end-to-end training fashion. Specifically, the depth of our network can be easily extended to hundreds of layers. Our Residual Attention Network outperforms state-of-the-art residual networks on CIFAR-10, CIFAR-100 and challenging ImageNet~\cite{deng2009imagenet} image classification dataset with significant reduction of computation (\textbf{69\%} forward FLOPs).

All of the aforementioned properties, which are challenging to achieve with previous approaches, are made possible with following contributions:

\noindent
(1) \textit{Stacked network structure}: Our Residual Attention Network is constructed by stacking multiple Attention Modules. The stacked structure is the basic application of mixed attention mechanism. Thus, different types of attention are able to be captured in different Attention Modules.

\noindent
(2) \textit{Attention Residual Learning}: Stacking Attention Modules directly would lead to the obvious performance drop. Therefore, we propose attention residual learning mechanism to optimize very deep Residual Attention Network with hundreds of layers. %Details

\noindent
(3) \textit{Bottom-up top-down feedforward attention}: Bottom-up top-down feedforward structure has been successfully applied to human pose estimation~\cite{newell2016stacked} and image segmentation~\cite{long2015fully,noh2015learning,badrinarayanan2015segnet}. We use such structure as part of Attention Module to add soft weights on features. This structure can mimic bottom-up fast feedforward process and top-down attention feedback in a single feedforward process which allows us to develop an end-to-end trainable network with top-down attention. The bottom-up top-down structure in our work differs from stacked hourglass network~\cite{newell2016stacked} in its intention of guiding feature learning.
\section{Related Work}

Evidence from human perception process~\cite{mnih2014recurrent} shows the importance of attention mechanism, which uses top information to guide bottom-up feedforward process. Recently, tentative efforts have been made towards applying attention into deep neural network. Deep Boltzmann Machine (DBM)~\cite{larochelle2010learning} contains top-down attention by its reconstruction process in the training stage. Attention mechanism has also been widely applied to recurrent neural networks (RNN) and long short term memory (LSTM) ~\cite{hochreiter1997long} to tackle sequential decision tasks~\cite{noh2015learning, srivastava2015training, larochelle2010learning, kim2016multimodal}. Top information is gathered sequentially and decides where to attend for the next feature learning steps.

Residual learning~\cite{resnet2016} is proposed to learn residual of identity mapping. This technique greatly increases the depth of feedforward neuron network. Similar to our work, ~\cite{noh2015learning, srivastava2015training, larochelle2010learning, kim2016multimodal} use residual learning with attention mechanism to benefit from residual learning. Two information sources (query and query context) are captured using attention mechanism to assist each other in their work. While in our work, a single information source (image) is split into two different ones and combined repeatedly. And residual learning is applied to alleviate the problem brought by repeated splitting and combining.

In image classification, top-down attention mechanism has been applied using different methods: sequential process, region proposal and control gates. Sequential process ~\cite{mnih2014recurrent,hendricks2015deep,xu2015show,gregor2015draw} models image classification as a sequential decision. Thus attention can be applied similarly with above. This formulation allows end-to-end optimization using RNN and LSTM and can capture different kinds of attention in a goal-driven way.

Region proposal~\cite{shrivastava2016contextual,dai2015convolutional,hariharan2014simultaneous,yang2015faceness} has been successfully adopted in image detection task. In image classification, an additional region proposal stage is added before feedforward classification. The proposed regions contain top information and are used for feature learning in the second stage. Unlike image detection whose region proposals rely on large amount of supervision, e.g. the ground truth bounding boxes or detailed segmentation masks~\cite{erhan2014scalable}, unsupervised learning~\cite{xiao2015application} is usually used to generate region proposals for image classification.

Control gates have been extensively used in LSTM.  In image classification with attention, control gates for neurones are updated with top information and have influence on the feedforward process during training~\cite{cao2015look,stollenga2014deep}. However, a new process, reinforcement learning~\cite{stollenga2014deep} or optimization~\cite{cao2015look} is involved during the training step. Highway Network~\cite{srivastava2015training} extends control gate to solve gradient degradation problem for deep convolutional neural network.

However, recent advances of image classification focus on training feedforward convolutional neural networks using ``very deep" structure~\cite{simonyan2014very,szegedy2015going,resnet2016}. The feedforward convolutional network mimics the bottom-up paths of human cortex. Various approaches have been proposed to further improve the discriminative ability of deep convolutional neural network. VGG~\cite{simonyan2014very}, Inception~\cite{szegedy2015going} and residual learning~\cite{resnet2016} are proposed to train very deep neural networks. Stochastic depth~\cite{huang2016deep}, Batch Normalization~\cite{BN2015} and Dropout~\cite{dropout2014} exploit regularization for convergence and avoiding overfitting and degradation.

Soft attention developed in recent work~\cite{chen2015attention, jaderberg2015spatial} can be trained end-to-end for convolutional network. Our Residual Attention Network incorporates the soft attention in fast developing feedforward network structure in an innovative way. Recent proposed spatial transformer module~\cite{jaderberg2015spatial} achieves state-of-the-art results on house number recognition task. A deep network module capturing top information is used to generate affine transformation. The affine transformation is applied to the input image to get attended region and then feed to another deep network module. The whole process can be trained end-to-end by using differentiable network layer which performs spatial transformation. Attention to scale~\cite{chen2015attention} uses soft attention as a scale selection mechanism and gets state-of-the-art results in image segmentation task.

The design of soft attention structure in our Residual Attention Network is inspired by recent development of localization oriented task, \ie segmentation~\cite{long2015fully,noh2015learning,badrinarayanan2015segnet} and human pose estimation~\cite{newell2016stacked}. These tasks motivate researchers to explore structure with fined-grained feature maps. The frameworks tend to cascade a bottom-up and a top-down structure. The bottom-up feedforward structure produces low resolution feature maps with strong semantic information. After that, a top-down network produces dense features to inference on each pixel. Skip connection~\cite{long2015fully} is employed between bottom and top feature maps and achieved state-of-the-art result on image segmentation. The recent stacked hourglass network~\cite{newell2016stacked} fuses information from multiple scales to predict human pose, and benefits from encoding both global and local information.

%-------------------------------------------------------------------------
\section{Residual Attention Network}

\begin{figure*}[t]
\setlength{\abovecaptionskip}{0pt}

\begin{center}
  \includegraphics[width=1.0\linewidth]{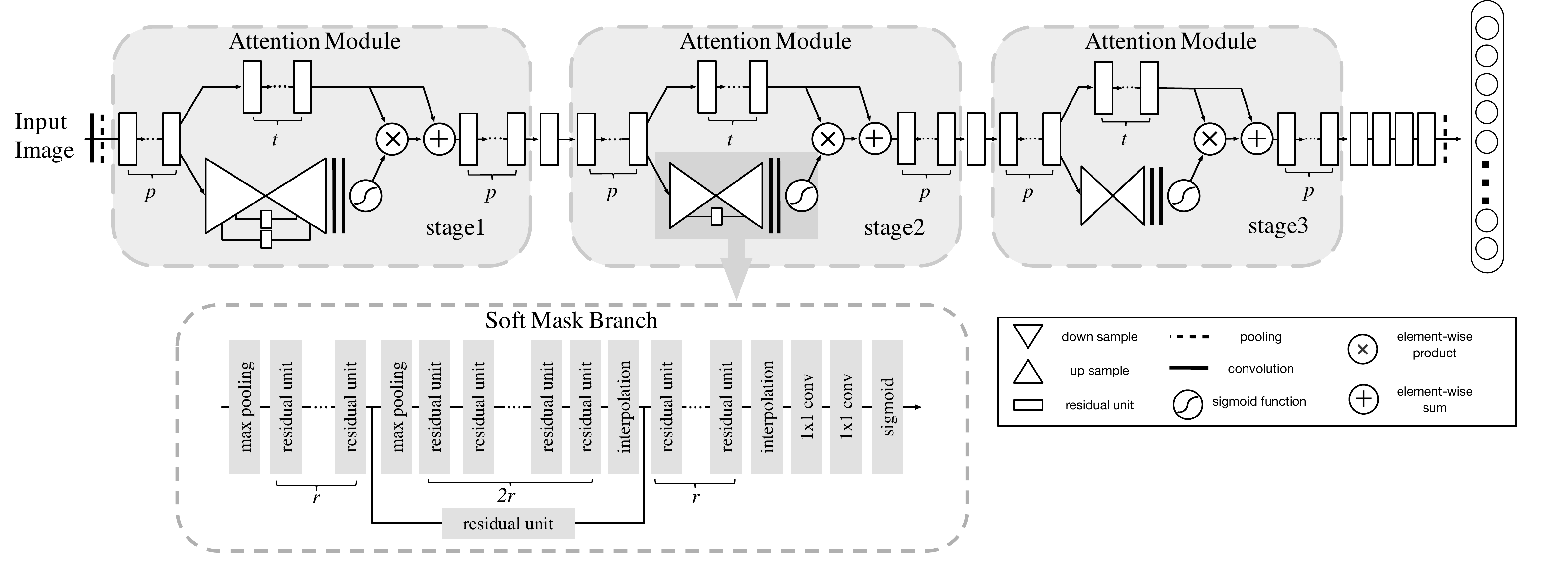}
\end{center}
   \caption{Example architecture of the proposed network for ImageNet. We use three hyper-parameters for the design of Attention Module: $p,t$ and $r$. The hyper-parameter $p$ denotes the number of pre-processing Residual Units before splitting into trunk branch and mask branch. $t$ denotes the number of Residual Units in trunk branch. $r$ denotes the number of Residual Units between adjacent pooling layer in the mask branch. In our experiments, we use the following hyper-parameters setting: $\{p=1$, $t=2$, $r=1\}$. The number of channels in the soft mask Residual Unit and corresponding trunk branches is the same.}

\label{fig:Attention}
\end{figure*}

Our Residual Attention Network is constructed by stacking multiple Attention Modules. Each Attention Module is divided into two branches: mask branch and trunk branch. The trunk branch performs feature processing and can be adapted to any state-of-the-art network structures.
In this work, we use pre-activation Residual Unit~\cite{he2016identity}, ResNeXt~\cite{resnext} and Inception~\cite{inception} as our Residual Attention Networks basic unit to construct Attention Module. Given trunk branch output $T(x)$ with input $x$, the mask branch uses bottom-up top-down structure~\cite{long2015fully, noh2015learning, badrinarayanan2015segnet, newell2016stacked} to learn same size mask $M(x)$ that softly weight output features $T(x)$. The bottom-up top-down structure mimics the fast feedforward and feedback attention process. The output mask is used as control gates for neurons of trunk branch similar to Highway Network~\cite{srivastava2015training}. The output of Attention Module $H$ is:
\begin{equation}
H_{i,c}(x)=M_{i,c}(x)*T_{i,c}(x)
\end{equation}
where i ranges over all spatial positions and $c\in \{1,...,C\}$ is the index of the channel. The whole structure can be trained end-to-end.

In Attention Modules, the attention mask can not only serve as a feature selector during forward inference, but also as a gradient update filter during back propagation. In the soft mask branch, the gradient of mask for input feature is:
\begin{equation}
\frac{\partial M(x, \theta)T(x,\phi)}{\partial \phi} = M(x, \theta)\frac{\partial T(x,\phi)}{\partial \phi}
\end{equation}
\noindent
where the $\theta$ are the mask branch parameters and the $\phi$ are the trunk branch parameters. This property makes Attention Modules robust to noisy labels. Mask branches can prevent wrong gradients (from noisy labels) to update trunk parameters. Experiment in Sec.\ref{para:noise} shows the robustness of our Residual Attention Network against noisy labels.

Instead of stacking Attention Modules in our design, a simple approach would be using a single network branch to generate soft weight mask, similar to spatial transformer layer~\cite{jaderberg2015spatial}. However, these methods have several drawbacks on challenging datasets such as ImageNet. First, images with clutter background, complex scenes, and large appearance variations need to be modeled by different types of attentions. In this case, features from different layers need to be modeled by different attention masks. Using a single mask branch would require exponential number of channels to capture all combinations of different factors. Second, a single Attention Module only modify the features once. If the modification fails on some parts of the image, the following network modules do not get a second chance.

The Residual Attention Network alleviates above problems. In Attention Module, each trunk branch has its own mask branch to learn attention that is specialized for its features. As shown in Fig.\ref{fig:motivation}, in hot air balloon images, blue color features from bottom layer have corresponding sky mask to eliminate background, while part features from top layer are refined by balloon instance mask. Besides, the incremental nature of stacked network structure can gradually refine attention for complex images.

\subsection{Attention Residual Learning}
However, naive stacking Attention Modules leads to the obvious performance drop. First, dot production with mask range from zero to one repeatedly will degrade the value of features in deep layers. Second, soft mask can potentially break good property of trunk branch, for example, the identical mapping of Residual Unit.

We propose attention residual learning to ease the above problems. Similar to ideas in residual learning, if soft mask unit can be constructed as identical mapping, the performances should be no worse than its counterpart without attention. Thus we modify output $H$ of Attention Module as
\begin{equation}
H_{i,c}(x)=(1+M_{i,c}(x))*F_{i,c}(x)
\end{equation}
$M(x)$ ranges from $[0,1]$, with $M(x)$ approximating 0, $H(x)$ will approximate original features $F(x)$. We call this method attention residual learning.
\\
\indent
Our stacked attention residual learning is different from residual learning. In the origin ResNet, residual learning is formulated as $H_{i,c}(x)= x + F_{i,c}(x)$, where $F_{i,c}(x)$ approximates the residual function. In our formulation, $F_{i,c}(x)$ indicates the features generated by deep convolutional networks. The key lies on our mask branches $M(x)$. They work as feature selectors which enhance good features and suppress noises from trunk features.
\\
\indent
In addition, stacking Attention Modules backs up attention residual learning by its incremental nature. Attention residual learning can keep good properties of original features, but also gives them the ability to bypass soft mask branch and forward to top layers to weaken mask branch's feature selection ability. Stacked Attention Modules can gradually refine the feature maps. As show in Fig.\ref{fig:motivation}, features become much clearer as depth going deeper. By using attention residual learning, increasing depth of the proposed Residual Attention Network can improve performance consistently. As shown in the experiment section, the depth of Residual Attention Network is increased up to 452 whose performance surpasses ResNet-1001 by a large margin on CIFAR dataset.

\subsection{Soft Mask Branch}
Following previous attention mechanism idea in DBN~\cite{larochelle2010learning}, our mask branch contains fast feed-forward sweep and top-down feedback steps. The former operation quickly collects global information of the whole image, the latter operation combines global information with original feature maps. In convolutional neural network, the two steps unfold into bottom-up top-down fully convolutional structure.

\begin{figure}[t]
\setlength{\belowcaptionskip}{-10pt}
\setlength{\abovecaptionskip}{0pt}
\begin{center}
   \includegraphics[width=1\linewidth]{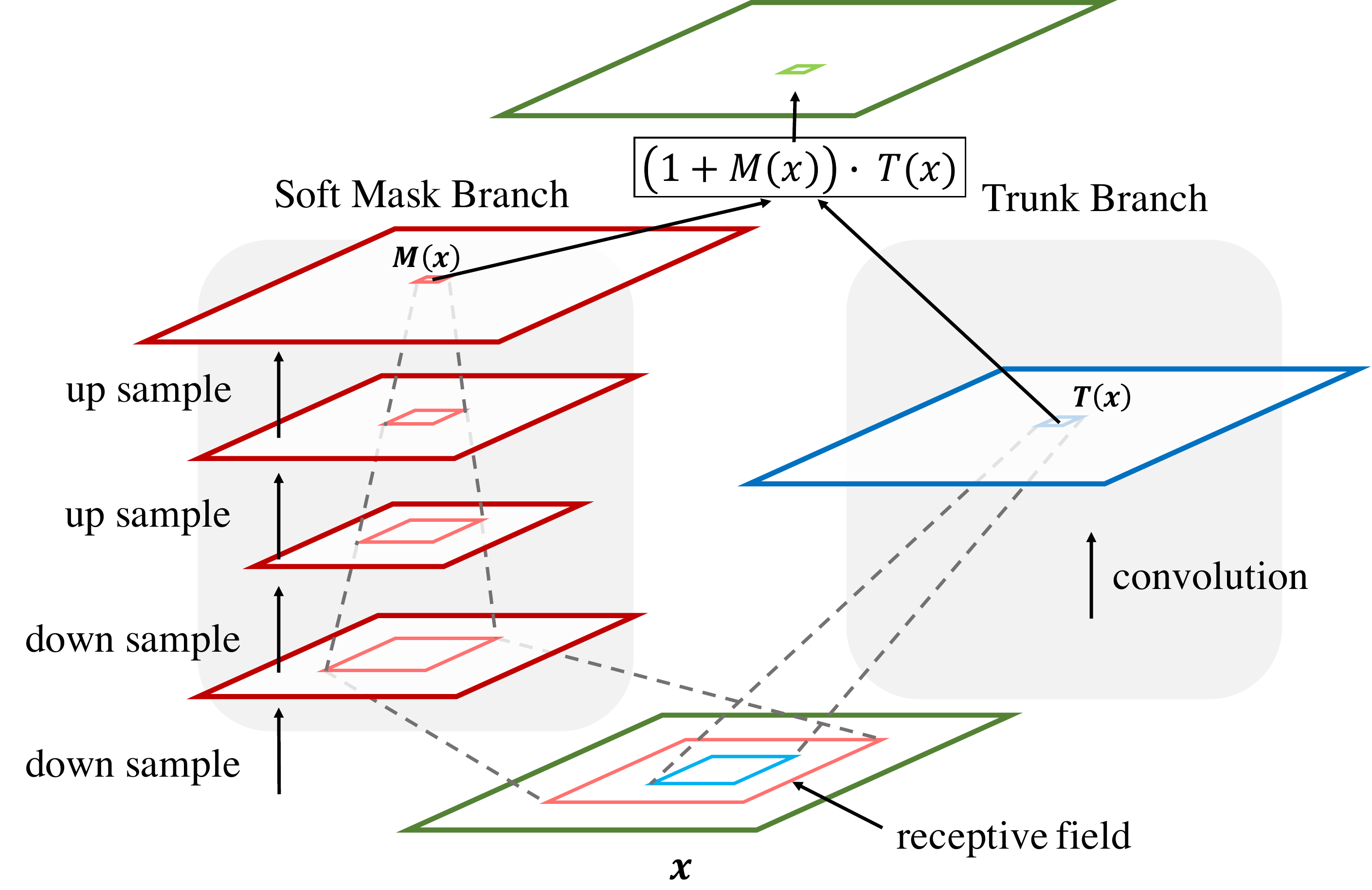}
\end{center}
   \caption{The receptive field comparison between mask branch and trunk branch.}
\label{fig:attentionunit}
\end{figure}

From input, max pooling are performed several times to increase the receptive field rapidly after a small number of Residual Units. After reaching the lowest resolution, the global information is then expanded by a symmetrical top-down architecture to guide input features in each position. Linear interpolation up sample the output after some Residual Units. The number of bilinear interpolation is the same as max pooling to keep the output size the same as the input feature map. Then a sigmoid layer normalizes the output range to $[0,1]$ after two consecutive $1\times 1$ convolution layers. We also added skip connections between bottom-up and top-down parts to capture information from different scales. The full module is illustrated in Fig.\ref{fig:Attention}.

The bottom-up top-down structure has been applied to image segmentation and human pose estimation. However, the difference between our structure and the previous one lies in its intention. Our mask branch aims at improving trunk branch features rather than solving a complex problem directly. Experiment in Sec.\ref{para:Comparison} is conducted to verify above arguments.
%Using additional classification supervision on mask branch directly leads to 0.5\% performance drop on CIFAR-10.

\subsection{Spatial Attention and Channel Attention}
In our work, attention provided by mask branch changes adaptably with trunk branch features. However, constrains to attention can still be added to mask branch by changing normalization step in activation function before soft mask output. We use three types of activation functions corresponding to mixed attention, channel attention and spatial attention. Mixed attention $f_{1}$ without additional restriction use simple sigmoid for each channel and spatial position. Channel attention $f_{2}$ performs $L2$ normalization within all channels for each spatial position to remove spatial information. Spatial attention $f_{3}$ performs normalization within feature map from each channel and then sigmoid to get soft mask related to spatial information only.
\begin{eqnarray}
&&f_{1}(x_{i,c}) = \frac{1}{1+ exp(-x_{i,c})}\\
&&f_{2}(x_{i,c}) = \frac{x_{i,c}}{\|x_{i}\|}\\
&&f_{3}(x_{i,c}) = \frac{1}{1+ exp(-(x_{i,c} - \text{mean}_c) / \text{std}_c)}
\end{eqnarray}
Where $i$ ranges over all spatial positions and $c$ ranges over all channels. $\text{mean}_c$ and $\text{std}_c$ denotes the mean value and standard deviation of feature map from $c$-th channel. $x_{i}$ denotes the feature vector at the $i$th spatial position.

\begin{table}\small
\setlength{\abovecaptionskip}{0pt}
\setlength{\belowcaptionskip}{-5pt}
\begin{center}
\begin{tabular}{c|c|c} \hline
Activation Function & Attention Type & Top-1 err. (\%) \\
\hline
$f_{1}(x)$ & Mixed Attention &\textbf{5.52}\\
\hline
$f_{2}(x)$  & Channel Attention &6.24\\
\hline
$f_{3}(x)$ & Spatial Attention &6.33\\
\hline
\end{tabular}
\end{center}
\caption{The test error (\%) on CIFAR-10 of Attention-56 network with different activation functions.}
\label{tab:activation_exp}
\end{table}

The experiment results are shown in Table~\ref{tab:activation_exp}, the mixed attention has the best performance. Previous works normally focus on only one type of attention, for example scale attention~\cite{chen2015attention} or spatial attention~\cite{jaderberg2015spatial}, which puts additional constrain on soft mask by weight sharing or normalization. However, as supported by our experiments, making attention change adaptively with features without additional constraint leads to the best performance.

\begin{table}\small
\setlength{\abovecaptionskip}{0pt}
\setlength{\belowcaptionskip}{-10pt}
\footnotesize
\begin{center}
\begin{tabular}{c|c|c|c} \hline

Layer &Output Size &Attention-56&Attention-92 \\
\hline
Conv1 & 112$\times$112 & \multicolumn{2}{|c}{$7\times 7$, 64, stride 2}  \\
\hline
Max pooling & 56$\times$56& \multicolumn{2}{|c}{$3\times 3$ stride 2}  \\
\hline
Residual Unit& 56$\times$56 & \multicolumn{2}{|c}{
$\left(
	\begin{matrix}
	1\times 1, 64 \\
	3\times 3, 64 \\
	1\times	1, 256
	\end{matrix}
\right)\times 1$
}  \\
\hline
Attention Module& 56$\times$56 & Attention $\times$1 & Attention $\times$1  \\
\hline
Residual Unit& 28$\times$28 & \multicolumn{2}{|c}{
$\left(
	\begin{matrix}
	1\times 1, 128 \\
	3\times 3, 128 \\
	1\times	1, 512
	\end{matrix}
\right)\times 1$
}  \\
\hline
Attention Module& 28$\times$28 & Attention $\times$1 & Attention $\times$2  \\
\hline
Residual Unit& 14$\times$14 & \multicolumn{2}{|c}{
$\left(
	\begin{matrix}
	1\times 1, 256 \\
	3\times 3, 256 \\
	1\times	1, 1024
	\end{matrix}
\right)\times 1$
}  \\
\hline
Attention Module& 14$\times$14 & Attention $\times$1 & Attention $\times$3  \\
\hline
Residual Unit& 7$\times$7 & \multicolumn{2}{|c}{
$\left(
	\begin{matrix}
	1\times 1, 512 \\
	3\times 3, 512 \\
	1\times	1, 2048
	\end{matrix}
\right)\times 3$
}  \\
\hline
Average pooling & 1$\times$1& \multicolumn{2}{|c}{$7\times 7$ stride 1}  \\
\hline
FC,Softmax & \multicolumn{3}{|c}{1000}  \\
\hline
\multicolumn{2}{c|}{params$\times 10^6$} & $31.9$ & $51.3$  \\
\hline
\multicolumn{2}{c|}{FLOPs$\times 10^9$} & $6.2$ &$10.4$  \\
\hline
\multicolumn{2}{c|}{Trunk depth} & $56 $ & $92$  \\
\hline

\end{tabular}
\end{center}
\caption{Residual Attention Network architecture details for ImageNet. Attention structure is described in Fig.~\ref{fig:Attention}.  We make the size of the smallest output map in each mask branch 7$\times$7 to be consistent with the smallest trunk output map size. Thus 3,2,1 max-pooling layers are used in mask branch with input size 56$\times$56, 28$\times$28, 14$\times$14 respectively.
The Attention Module is built by pre-activation Residual Unit~\cite{he2016identity} with the number of channels in each stage is the same as ResNet~\cite{resnet2016}.
}
\label{tab:attention_structure}
\end{table}

%-------------------------------------------------------------------------
\section{Experiments}
In this section, we evaluate the performance of proposed Residual Attention Network on a series of benchmark datasets including CIFAR-10, CIFAR-100~\cite{krizhevsky2009learning}, and ImageNet~\cite{deng2009imagenet}.
Our experiments contain two parts. In the first part, we analyze the effectiveness of each component in the Residual Attention Network including attention residual learning mechanism and different architectures of soft mask branch in the Attention Module.
After that, we explore the noise resistance property. Given limited computation resources, we choose CIFAR-10 and CIFAR-100 dataset to conduct these experiments. Finally, we compare our network with state-of-the-art results in CIFAR dataset.
In the second part, we replace the Residual Unit with Inception Module and ResNeXt to demonstrate our Residual Attention Network surpasses origin networks both in parameter efficiency and final performance.
We also compare image classification performance with state-of-the-art ResNet and Inception on ImageNet dataset.

\subsection{CIFAR and Analysis}

\paragraph{Implementation.}
\phantomsection
\label{para:imple}
The CIFAR-10 and CIFAR-100 datasets consist of $60,000$ $32\times32$ color images of $10$ and $100$ classes respectively, with $50,000$ training images and $10,000$ test images.
The broadly applied state-of-the-art network structure ResNet is used as baseline method.
To conduct fair comparison, we keep most of the settings same as ResNet paper~\cite{resnet2016}.
The image is padded by 4 pixels on each side, filled with $0$ value resulting in $40\times40$ image. A $32\times32$ crop is randomly sampled from an image or its horizontal flip, with the per-pixel RGB mean value subtracted.
We adopt the same weight initialization method following previous study~\cite{prelu2015} and train Residual Attention Network using nesterov SGD with a mini-batch size of 64.
We use a weight decay of $0.0001$ with a momentum of $0.9$ and set the initial learning rate to 0.1. The learning rate is divided by 10 at $64$k and $96$k iterations. We terminate training at $160$k iterations.

The overall network architecture and the hyper parameters setting are described in Fig.\ref{fig:Attention}.
The network consists of 3 stages and similar to ResNet~\cite{resnet2016}, equal number of Attention Modules are stacked in each stage.
Additionally, we add two Residual Units at each stage. The number of weighted layers in trunk branch is 36$m$+20 where $m$ is the number of Attention Module in one stage.
We use original $32\times32$ image for testing.

\paragraph{Attention Residual Learning.}

In this experiment, we evaluate the effectiveness of attention residual learning mechanism.
Since the notion of attention residual learning (ARL) is new, no suitable previous methods are comparable therefore we use ``naive attention learning'' (NAL) as baseline.
Specifically, ``naive attention learning'' uses Attention Module where features are directly dot product by soft mask without attention residual learning.
%
% Add a small figure here.
We set the number of Attention Module in each stage $m$ = \{1, 2, 3, 4\}. For Attention Module, this leads to Attention-56 (named by trunk layer depth), Attention-92, Attention-128 and Attention-164 respectively.
\begin{table}\small
\setlength{\abovecaptionskip}{0pt}
\setlength{\belowcaptionskip}{-10pt}
\begin{center}
\begin{tabular}{c|c|c} \hline
 Network & ARL (Top-1 err. \%) & NAL (Top-1 err.\%)\\
\hline
Attention-56 &\textbf{5.52} & 5.89\\
\hline
Attention-92 &\textbf{4.99} & 5.35\\
\hline
Attention-128 &\textbf{4.44} & 5.57\\
\hline
Attention-164 &\textbf{4.31} & 7.18\\
\hline
\end{tabular}
\end{center}
\caption{
Classification error (\%) on CIAFR-10.}
\label{tab:learning}
\end{table}

We train these networks using different mechanisms and summarize the results in the Table~\ref{tab:learning}.
As shown in Table~\ref{tab:learning}, the networks trained using attention residual learning technique consistently outperform the networks trained with baseline method which proves the effectiveness of our method. 
The performance increases with the number of Attention Module when applying attention residual learning. In contrast, the performance of networks trained with ``naive attention learning'' method suffers obvious degradation with increased number of Attention Module.

\begin{figure}[t]
\setlength{\abovecaptionskip}{0pt}
\setlength{\belowcaptionskip}{-15pt}
\begin{center}
%\fbox{\rule{0pt}{2in} \rule{0.9\linewidth}{0pt}}%
  \includegraphics[width=1\linewidth]{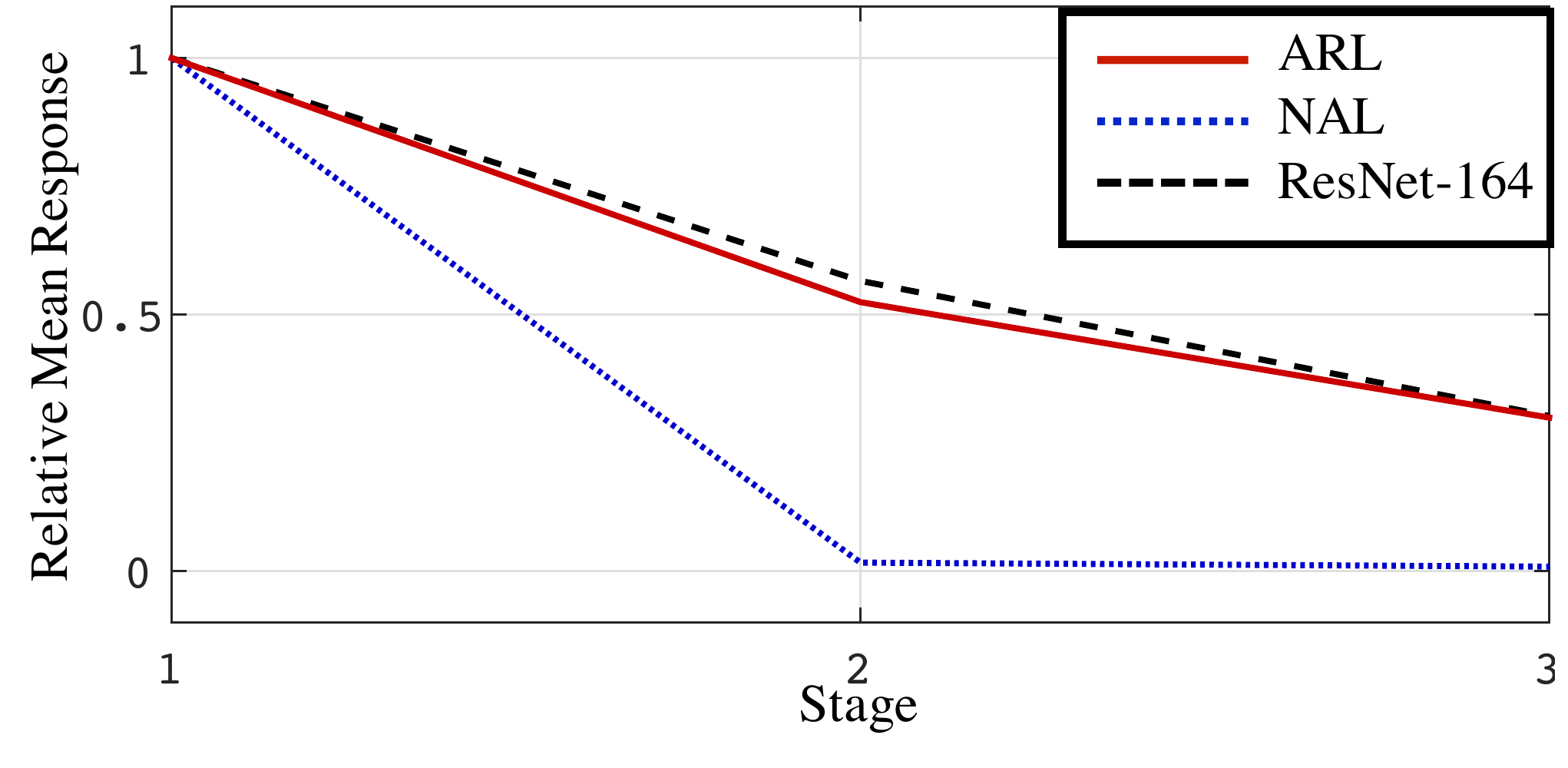}
\end{center}
   \caption{The mean absolute response of output features in each stage. }
\label{fig:mean_response}
\end{figure}
To understand the benefit of attention residual learning, we calculate mean absolute response value of output layers for each stage. We use Attention-164 to conduct this experiment.
As shown in the Fig.~\ref{fig:mean_response}, the response generated by the network trained using naive attention learning quickly vanishes in the stage 2 after four Attention Modules compared with network trained using attention residual learning.
The Attention Module is designed to suppress noise while keeping useful information by applying dot product between feature and soft mask. However, repeated dot product will lead to severe degradation of both useful and useless information in this process.
The attention residual learning can relieve signal attenuation using identical mapping, which enhances the feature contrast.
Therefore, it gains benefits from noise reduction without significant information loss, which makes optimization much easier while improving the discrimination of represented features.
In the rest of the experiments, we apply this technique to train our networks.

%----------------------------------------------------------------------------------
\paragraph{Comparison of different mask structures.}
\label{para:Comparison}
We conduct experiments to validate the effectiveness of encoder-decoder structure by comparing with local convolutions without any down sampling or up sampling. The local convolutions soft mask consists of three Residual Units using the same number of FLOPs.
The Attention-56 is used to construct Attention-Encoder-Decoder-56 and Attention-Local-Conv-56 respectively.
Results are shown in Table~\ref{tab:local_global_attention}.
The Attention-Encoder-Decoder-56 network achieves lower test error $5.52\%$ compared with Attention-Local-Conv-56 network $6.48\%$ with a considerable margin $0.94\%$. The result suggests that the soft attention optimization process will benefit from multi-scale information.

\begin{table}[h]\small
\setlength{\abovecaptionskip}{0pt}
\begin{center}
\begin{tabular}{c|c|c} \hline
Mask Type  & Attention Type &Top-1 err. (\%) \\
\hline
Local Convolutions & Local Attention &6.48 \\
\hline
Encoder and Decoder  & Mixed Attention &\textbf{5.52}\\
\hline
\end{tabular}
\end{center}
\caption{Test error (\%) on CIFAR-10 using different mask structures.}
\label{tab:local_global_attention}
\end{table}

\paragraph{Noisy Label Robustness.}
\label{para:noise}

\begin{table}\small
\setlength{\abovecaptionskip}{0pt}
\setlength{\belowcaptionskip}{-10pt}
\begin{center}
\begin{tabular}{c|c|c} \hline
Noise Level &ResNet-164 err. (\%) & Attention-92 err. (\%) \\
\hline
10\% &5.93 &5.15\\
\hline
30\% &6.61 &5.79\\
\hline
50\% &8.35 &7.27\\
\hline
70\% &17.21 &15.75\\
\hline
\end{tabular}
\end{center}
\caption{Test error (\%) on CIFAR-10 with label noises.}
\label{tab:noise_label}
\end{table}
In this experiment, we show our Residual Attention Network enjoys noise resistant property on CIFAR-10 dataset following the setting of paper~\cite{sukhbaatar2014training}.
The confusion matrix $Q$ in our experiment is set as follows:
\begin{equation}
Q =
\left(
\begin{matrix}
r & \frac{1-r}{9} &\cdots &\frac{1-r}{9} \\
\frac{1-r}{9} &r  &\cdots &\frac{1-r}{9} \\
\vdots & \vdots & \ddots & \vdots \\
\frac{1-r}{9} & \frac{1-r}{9} &\cdots &r \\
\end{matrix}
\right)_{10\times 10}
\end{equation}

\noindent
where $r$ denotes the clean label ratio for the whole dataset.

We compare ResNet-164 network with Attention-92 network under different noise levels.
The Table~\ref{tab:noise_label} shows the results.
The test error of Attention-92 network is significantly lower than ResNet-164 network with the same noise level.
In addition, when we increase the ratio of noise, test error of Attenion-92 declines slowly compared with ResNet-164 network.
These results suggest that our Residual Attention Network can perform well even trained with high level noise data.
%
%The encode-decode structure can fast feedforward the whole image and obtain the global and local information of image.
%
When the label is noisy, the corresponding mask can prevent gradient caused by label error to update trunk branch parameters in the network.
In this way, only the trunk branch is learning the wrong supervision information and soft mask branch masks the wrong label.

\paragraph{Comparisons with state-of-the-art methods.}
\begin{table}\small
\setlength{\abovecaptionskip}{0pt}
\setlength{\belowcaptionskip}{-5pt}
\begin{center}
\resizebox{\linewidth}{!}{%
\begin{tabular}{c|c|c|c}
	\hline
   	Network& params$\times 10^6$ & CIFAR-10  &  CIFAR-100 \\
	\hline
	ResNet-164~\cite{he2016identity}  & 1.7   & 5.46  & 24.33 \\
	ResNet-1001~\cite{he2016identity} & 10.3   & 4.64  & 22.71 \\
	\hline
	WRN-16-8~\cite{zagoruyko2016wide} & 11.0   & 4.81  & 22.07 \\
	WRN-28-10~\cite{zagoruyko2016wide} & 36.5   & 4.17  & 20.50 \\
	\hline
	Attention-92 & 1.9 & 4.99 & 21.71 \\
	Attention-236 & 5.1 & 4.14 & 21.16 \\
	Attention-452$\dag$ & 8.6 & \textbf{3.90}  & \textbf{20.45}\\
	\hline
\end{tabular}
}
\end{center}
\caption{Comparisons with state-of-the-art methods on CIFAR-10/100. $\dag$: the Attention-452 consists of Attention Module with hyper-parameters setting: $\{p=2$, $t=4$, $r=3\}$ and 6 Attention Modules per stage. }
\label{tab:cifar_results}
\end{table}

We compare our Residual Attention Network with state-of-the-art methods including ResNet~\cite{he2016identity} and Wide ResNet~\cite{zagoruyko2016wide} on CIFAR-10 and CIFAR-100 datasets.
The results are shown in Table~\ref{tab:cifar_results}.
Our Attention-452 outperforms all the baseline methods on CIFAR-10 and CIFAR-100 datasets.
Note that Attention-92 network achieves $4.99\%$ test error on CIFAR-10 and $21.71\%$ test error on CIFAR-100 compared with $5.46\%$ and $24.33\%$ test error on CIFAR-10 and CIFAR-100 for ResNet-164 network under similar parameter size.
In addition, Attention-236 outperforms ResNet-1001 using only half of the parameters. It suggests that our Attention Module and attention residual learning scheme can effectively reduce the number of parameters in the network while improving the classification performance.
%
%It worth to mention that, our method is complementary with other state-of-the-art methods which focus on regularization and can achieve better results by applying these advanced techniques.

\subsection{ImageNet Classification}

In this section, we conduct experiments using ImageNet LSVRC $2012$ dataset~\cite{deng2009imagenet}, which contains $1,000$ classes with $1.2$ million training images, $50,000$ validation images, and $100,000$ test images.
The evaluation is measured on the non-blacklist images of the ImageNet LSVRC $2012$ validation set.
We use Attention-56 and Attention-92 to conduct the experiments. The network structures and hyper parameters can be found in the Table~\ref{tab:attention_structure}.

\paragraph{Implementation.}
Our implementation generally follows the practice in the previous study~\cite{krizhevsky2012imagenet}.
We apply scale and aspect ratio augmentation~\cite{szegedy2015going} to the original image.
A $224\times 224$ crop is randomly sampled from an augment image or its horizontal flip, with the per-pixel RGB scale to $[0,1]$ and mean value subtracted and standard variance divided. We adopt standard color augmentation~\cite{krizhevsky2012imagenet}.
The network is trained using SGD with a momentum of $0.9$.
We set initial learning rate to 0.1. The learning rate is divided by 10 at $200$k, $400$k, $500$k iterations. We terminate training at $530$k iterations.

\paragraph{Mask Influence.}

\begin{table*}\small
\setlength{\abovecaptionskip}{0pt}
\setlength{\belowcaptionskip}{-10pt}
\begin{center}
%\resizebox{\linewidth}{!}{%
\begin{tabular}{c|c|c|c|c|c} \hline
Network & params$\times 10^6$ &FLOPs$\times 10^9$ & Test Size &Top-1 err. (\%) &Top-5 err. (\%) \\
\hline
ResNet-152~\cite{resnet2016}  &60.2 &11.3 &$224\times224$&22.16 &6.16\\
\hline
Attention-56 &31.9 &6.3 &$224\times224$&\textbf{21.76} &\textbf{5.9} \\
\hline
\hline
ResNeXt-101 ~\cite{resnext}&44.5 & 7.8&$224\times224$    &21.2 &5.6 \\
\hline
AttentionNeXt-56 &31.9 & 6.3&$224\times224$  &\textbf{21.2} &\textbf{5.6} \\
\hline
\hline
Inception-ResNet-v1~\cite{inception} &- &-&$299\times299$&21.3 &5.5 \\
\hline
AttentionInception-56 &31.9 & 6.3 &$299\times299$ &\textbf{20.36} &\textbf{5.29} \\
\hline
\hline
ResNet-200~\cite{he2016identity} &64.7 &15.0 &$320\times320$ &20.1  &4.8 \\
\hline
{Inception-ResNet-v2} &- &- &$299\times299$ &19.9  &4.9 \\
\hline
Attention-92 &51.3  & 10.4&$320\times320$ &\textbf{19.5 }  &\textbf{4.8} \\
\hline
\end{tabular}
%}
\end{center}
	\caption{Single crop validation error on ImageNet.
}
\label{tab:single_crop_validation_error}
\end{table*}

In this experiment, we explore the efficiency of proposed Residual Attention Network.
We compare Attention-56 with ResNet-152~\cite{resnet2016}.
The ResNet-152 has 50 trunk Residual Units and 60.2$\times 10^6$ parameters compared with 18 trunk Residual Units and 31.9$\times 10^6$ parameters in Attention-56.
We evaluate our model using single crop scheme on the ImageNet validation set and show results in Table~\ref{tab:single_crop_validation_error}.
The Attention-56 network outperforms ResNet-152 by a large margin with a $0.4\%$ reduction on top-1 error and a $0.26\%$ reduction on top-5 error.
More importantly, Attention-56 network achieves better performance with only 52\% parameters and 56\% FLOPs compared with ResNet-152, which suggests that the proposed attention mechanism can significantly improve network performance while reducing the model complexity.

\paragraph{Different Basic Units.}
In this experiment, we show Residual Attention Network can generalize well using different basic unit. We apply three popular basic units: Residual Unit, ResNeXt~\cite{resnext}, and Inception~\cite{inception} to construct our Residual Attention Networks. To keep the number of parameters and FLOPs in the same scale, we simplify the Inception. Results are shown in Table~\ref{tab:single_crop_validation_error}.

%
%\begin{figure}[t]
%\setlength{\abovecaptionskip}{0pt}
%\setlength{\belowcaptionskip}{-10pt}
%\begin{center}
%\fbox{\rule{0pt}{2in} \rule{0.9\linewidth}{0pt}}%
%  \includegraphics[width=1\linewidth]{images/inception.png}
  %\includegraphics{images/whole_net.eps}
%\end{center}
%   \caption{The simple inception module stucture. The hyper-parameter $c$ denotes the number of channel in one stage. In this experiment, we choose $\{256, 512, 1024, 2048\}$ at feature map $\{56\times56, 28\times28, 14\times14, 7\times7\}$.}
%\label{fig:inception}
%\end{figure}
%
When the basic unit is ResNeXt, the AttentionNeXt-56 network performance is the same as ResNeXt-101 while the parameters and FLOPs are significantly fewer than ResNeXt-101.
For Inception, The AttentionIncepiton-56 outperforms Inception-ResNet-v1~\cite{inception} by a margin with a 0.94\% reduction on top-1 error and a 0.21\% reduction on top-5 error.
The results show that our method can be applied on different network structures.

\paragraph{Comparisons with State-of-the-art Methods.}

%\begin{table}
%\setlength{\belowcaptionskip}{-10pt}
%\begin{center}
%\resizebox{\linewidth}{!}{%
%\begin{tabular}{c|c|c|c|c} \hline
%Network &param/M & FLOPs$\times 10^9$ &top-1 err. &top-5 err.\\
%\hline
%{ResNet-200}~\cite{he2016identity} &64.7 &15.0  &20.1  &4.8 \\
%\hline
%{Inception-ResNet-v2} &- &-  &19.9  &4.9 \\
%\hline
%Attention-92 &51.3  & 10.4 &\textbf{19.5 }  &\textbf{4.8} \\
%\hline
%\end{tabular}
%}
%\end{center}
 %\caption{
%Comparisons of single crop error on the ILSVRC 2012 validation set. In order to compare fairly, we also test our Attention Network on a %single 320$\times$320 crop.}
%\label{tab:imagenet_result}
%\end{table}

We compare our Attention-92 evaluated using single crop on the ILSVRC 2012 validation set with state-of-the-art algorithms.
Table~\ref{tab:single_crop_validation_error} shows the results.
Our Attention-92 outperforms ResNet-200 with a large margin. The reduction on top-1 error is $0.6\%$.
Note that the ResNet-200 network contains $32\%$ more parameters than Attention-92.
The computational complexity of Attention-92 shown in the Table~\ref{tab:single_crop_validation_error} suggests that our network reduces nearly half training time comparing with ResNet-200 by adding attention mechanism and reducing trunk depth.
Above results suggest that our model enjoys high efficiency and good performance.

%Our architecture is parallel to major structure of original network, which is friendly to parallel computation. (3) Stacked Attention Module on $14\times14$ feature map gains $1.3\%$ improvement, contrast to the one of single unit, benefits from more Attention Module.

%Note that we test a single 320$\times$320 crop from short side of 320, which is consistent with ResNet-200[].
%Although our Attention-80 has significantly computation complexity than pre-activation ResNet-200 [](15.0$\times 10^9$), Our Attention-80 has achieved top-1 error rate of 20.3\%, which is 0.4\% lower than the baseline ResNet-200.

\section{Discussion}
We propose a Residual Attention Network which stacks multiple Attention Modules. The benefits of our network are in two folds: it can capture mixed attention and is an extensible convolutional neural network. The first benefit lies in that different Attention Modules capture different types of attention to guide feature learning. Our experiments on the forms of activation function also validate this point: free form mixed attention will have better performance than constrained (including single) attention. The second benefit comes from encoding top-down attention mechanism into bottom-up top-down feedforward convolutional structure in each Attention Module. Thus, the basic Attention Modules can be combined to form larger network structure. Moreover, residual attention learning allows training very deep Residual Attention Network. The performance of our model surpasses state-of-the-art image classification methods, \ie ResNet on CIFAR-10 (3.90\% error), CIFAR-100 (20.67\% error), and challenging ImageNet dataset (0.6\% top-1 accuracy improvement) with only $46\%$ trunk depth and $69\%$ forward FLOPs (comparing with ResNet-200). In the future, we will exploit different applications of deep Residual Attention Network such as detection and segmentation to better explore mixed attention mechanism for specific tasks.

{\small
\bibliographystyle{ieee}
\bibliography{attention-net_camera_ready_wf}
}

\end{document}